\documentclass{article}
\usepackage[utf8]{inputenc}

\usepackage{amsthm}
\usepackage{graphicx}
\usepackage{amsmath}
\usepackage{amssymb}
\usepackage{hyperref}

\begin{document}

\title{A Roadmap for End-to-End Robust Alignment}
\author{Lê Nguyên Hoang}
\maketitle




\begin{abstract}
As algorithms are becoming more and more data-driven, the greatest lever we have left to make them robustly beneficial to mankind lies in the design of their objective functions.
{\it Robust alignment} aims to address this design problem.
Arguably, the growing importance of social medias' recommender systems makes it an urgent problem, for instance to adequately automate hate speech moderation.
In this paper, we propose a preliminary research program for robust alignment.
This roadmap aims at decomposing the end-to-end alignment problem into numerous more tractable subproblems.
We hope that each subproblem is sufficiently orthogonal to others to be tackled independently, and that combining the solutions to all such subproblems may yield a solution to alignment.
\end{abstract}

\section{Introduction}

As they are becoming more capable and ubiquitous, algorithms are raising numerous concerns, including fairness, privacy, filter bubbles, addiction, job displacement or even existential risks \cite{russell2015,tegmark2017}. It has been argued that aligning the goals of algorithmic systems with humans' preferences would be an efficient way to make them reliably beneficial and to avoid potential catastrophic risks \cite{russell2019}. In fact, given the global influence of today's large-scale recommender systems \cite{kramer2014}, alignment has already become urgent. Indeed, such systems currently do not seem to be robustly beneficial, since they seem to recommend unscientific claims to hundreds of millions of users on important topic like climate change \cite{allgaier19}, and to be favoring the radicalization of many users \cite{ROWAM20}.

Crucially, in practice, the alignment of algorithms needs to be made {\it robust}. This means that it ought to be resilient to algorithmic bugs, flawed world models \cite{ha2018}, reward hacking \cite{amodei2016}, biased data \cite{mehrabi2019}, unforeseen side effects \cite{BABBC+18}, distributional shift \cite{SOFLN+19}, adversarial attacks \cite{GSS15} and moral uncertainty \cite{macaskill2014}. Robust alignment seems crucial in complex environments with diverging human preferences, like hate speech algorithmic moderation on social medias.


Unfortunately, robust alignment has been argued to be an extremely difficult problem \cite{bostrom2014,FGHLH+2020}. To address it, there have been a few attempts to list key research directions towards robust alignment \cite{soares2015,soares2017}. This paper aims to contribute to this line of work by outlining in a structured and compelling manner the main challenges posed by robust alignment. Given that the robustness of a system is often limited by its weakest component, robust alignment demands that we consider our algorithmic systems in their entirety. This motivates the proposal of a {\it roadmap} for {\it robust end-to-end alignment}, from data collection to algorithmic output.

While much of our proposal is speculative, we believe that several of the ideas presented here will be critical for the safety and alignment of algorithms. More importantly, we hope that this will be a useful roadmap to better estimate how they can best contribute to the effort. Given the complexity of the problem, our roadmap here will likely be full of gaps and false good ideas. Our proposal is nowhere near definite or perfect. Rather, we aim at presenting a sufficiently good starting point for others to build upon.

\section{The Roadmap}

\begin{figure*}[h!]
\centering
\includegraphics[width=\textwidth]{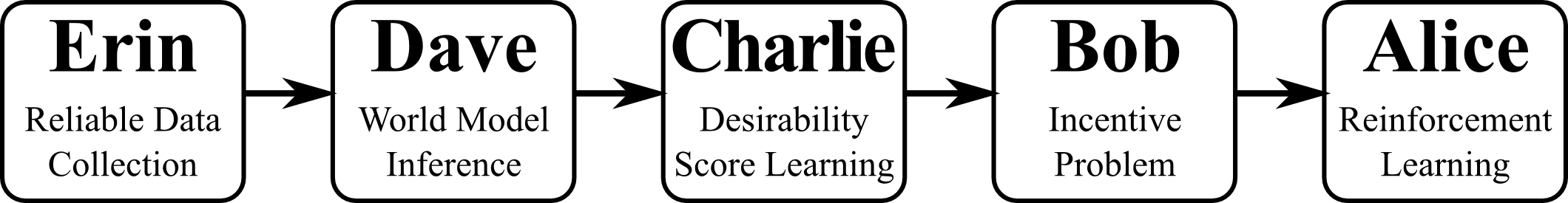}
\caption{We decompose the robust alignment problem into 5 key steps: data collection, world model inference, desirability learning, incentive design and reinforcement learning.}
\label{fig:roadmap}
\end{figure*}

Our roadmap consists of identifying key steps to robust alignment. For the sake of exposition, these steps will be personified by 5 characters, called Alice, Bob, Charlie, Dave and Erin. Roughly speaking, Erin will be collecting data from the world, Dave will use these data to infer the likely states of the world, Charlie will compute the desirability of the likely states of the world, Bob will derive incentive-compatible rewards to motivate Alice to take the right decision, and Alice will optimize decision-making. This decomposition is graphically represented in Figure \ref{fig:roadmap}.

Evidently, Alice, Bob, Charlie, Dave and Erin need not be 5 different algorithms. Typically, it may be much more computationally efficient to merge Charlie and Dave. Nevertheless, at least for pedagogical reasons, it seems useful to first dissociate the different roles that these components have.

In the sequel, we shall further detail the challenges posed by each of the 5 algorithms. We shall also argue that, for robustness and scalability reasons, these algorithms will need to be further divided into many more algorithms. We will see that this raises additional challenges. We shall also make a few non-technical remarks, before concluding.

\section{Alice's Reinforcement learning}

Today's most promising framework for large-scale algorithms seems to be {\it reinforcement learning} \cite{SuttonBarto1998,hutter2004}. In reinforcement learning, at each time $t$, the algorithm observes some state of the world $s_t$. Depending on its inner parameters $\theta_t$, it then takes (possibly randomly) some action $a_t$.

The decision $a_t$ then influences the next state and turns it into $s_{t+1}$. The transition from $s_t$ to $s_{t+1}$ given action $a_t$ is usually assumed to be nondeterministic. In any case, the algorithm then receives a reward $R_{t+1}$. The internal parameters $\theta_t$ of the algorithm may then be updated into $\theta_{t+1}$, depending on previous parameters $\theta_t$, action $a_t$, state $s_{t+1}$ and reward $R_{t+1}$.

Note that this is a very general framework. In fact, we humans are arguably (at least partially) subject to this framework. At any point in time, we observe new data $s_t$ that informs us about the world. Using an inner model of the world $\theta_t$, we then infer what the world probably is like, which motivates us to take some action $a_t$. This may affect what likely next data $s_{t+1}$ will be observed, and may be accompanied with a rewarding (or painful) feeling $R_{t+1}$, which will motivate us to update our inner model of the world $\theta_t$ into $\theta_{t+1}$.


Let us call Alice the algorithm in charge of performing this reinforcement learning reasoning. Alice can thus be viewed as an algorithm, which inputs observed states $s_t$ and rewards $R_t$, and undertakes actions $a_t$ so as to typically maximize some discounted sum of expected future rewards.

In the case of large-scale algorithms, such actions $a_t$ are mostly of the form of messages sent through the Internet. This may sound benign. But it is not. The YouTube recommender system might suggest billions of antivax videos, causing a major decrease of vaccination and an uprise of deadly diseases.
On a brighter note, if an algorithm now promotes convincing eco-friendly messages every day to billions of people, the public opinion on climate change may greatly change. Similarly, thoughtful compassion and cognitive empathy could be formidably enhanced by a robustly aligned recommender algorithm.

Note that, as opposed to all other components, in some sense, Alice is the real danger. Indeed, in our framework, she is the only one that really undertakes actions. More precisely, only her actions will be unconstrained (although others highly influence her decision-making and are thus critical as well). It is thus of the utmost importance that Alice be well-designed.

Some past work \cite{orseau2016,mhamdi2017} proposed to restrict the learning capabilities of Alice to provide provable desirable properties. Typically, they proposed to allow only a subclass of learning algorithms, i.e. of update rules of $\theta_{t+1}$ as a function of $(\theta_t, a_t, s_{t+1},R_{t+1})$. However, such restrictions might be too costly, and thus may not be adopted. Yet a safety measure will be useful only if the most powerful algorithms are {\it all} subject to the safety measure. As a result, an effective safety measure cannot hinder too much the performance of the algorithms. In other words, {\it there are constraints on the safety constraints that can be imposed}. This is what makes the research to make algorithms safe so challenging.

To design safe reinforcement learning algorithms, it seems more relevant to guarantee that they will do sufficient planning to avoid dangerous actions. One interesting idea by \cite{amodei2016} is {\it model lookahead}. This essentially corresponds to Alice simulating many likely scenarii before undertaking any action. More generally, Alice faces a {\it safe exploration problem}.

Another key feature of safe reinforcement learning algorithms is their ability to deal adequately with uncertainty. While recent results advanced the capability of algorithms to verify the robustness of neural networks \cite{DSGMK18,WWRHK20}, we should not expect algorithms to be correct all the time. Just like humans, algorithms will likely be sometime wrong. But then, even if an algorithm is right 99.9999\% of the time, it will still be wrong one time out of a million. Yet, algorithms like recommender systems or autonomous cars take billions of decisions every day. In such cases, thousands of algorithmic decisions may be unboundedly wrong every day.

This problem can become even more worrysome if we take into account the fact that hackers may attempt to take advantage of the algorithms' deficiencies. Such hackers may typically submit only data that corresponds to cases where the algorithms are wrong. This is known as {\it evasion attacks} \cite{lowd2005,su2017,gilmer2018}. To avoid evasion attacks, it is crucial for an algorithm to never be unboundedly wrong, e.g. by reliably measuring its own confidence in its decisions and to ask for help in cases of great uncertainty \cite{HDAR17}.

Now, even if Alice is well-designed, she will only be an effective optimization algorithm. Unfortunately, this is no guarantee of safety or alignment. Typically, because of humans' well-known confirmation bias \cite{haidt2012}, a watch-time maximization YouTube recommender algorithm may amplify filter bubbles, which may lead to worldwide geopolitical tensions. Both misaligned and unaligned algorithms will likely cause very undesirable {\it side effects} \cite{bostrom2014,everitt2018}.

To make sure that Alice will want to behave as we want her to, it seems critical to at least design appropriately the reward $R_{t+1}$. This is similar to the way children are taught to behave. We do so by punishing them when the sequence $(s_t,a_t,s_{t+1})$ is undesirable, and by rewarding them when the sequence $(s_t,a_t,s_{t+1})$ is desirable. In particular, rewards $R_t$ are Alice's incentives. They will determine her decision-making. Unfortunately, determining the adequate rewards $R_t$ to be given to Alice is an extremely difficult problem. It is, in fact, the key to robust alignment. Our roadmap to solve it identifies 4 key steps incarnated by Erin, Dave, Charlie and Bob.

\section{Erin's data collection problem}

Evidently, data are critical to do good. Indeed, even the most brilliant mind will be unable to know anything about the world if it does not have any data from that world. Now, much data is already available on the Internet, especially on a platform like YouTube. However, it is important to take into account the fact that the data on the Internet is not always fully reliable. It may be full of fake news, fraudulent entries, misleading videos, hacked posts and corrupted files.

It may then be relevant to invest in more reliable and relevant data collection. This would be Erin's job. Typically, Erin may want to collect economic metrics to better assess needs. Recently, it has been shown that satellite images combined with deep learning allow to compute all sorts of useful economic indicators \cite{jean2016}, including poverty risks and agricultural productivity. The increased use of sensors may further enable us to improve life standards, especially in developing countries.

To guarantee the reliability of such data, cryptographic and distributed computing solutions are likely to be useful as well, as they already are on the web. Electronic signatures may eventually be the key to distinguish reliable data from {\it deep fakes} \cite{JoPark2019,NKH2019}, while distributed computing, combined with recent Byzantine-resilient consensus algorithms like Blockchain \cite{nakamoto2008}, Hashgraph \cite{baird2016} or AT-2 \cite{GKMPS19}, could guarantee the reliable storage and traceability of critical information.

Note though that such data collection mechanisms could pose major privacy issues. It is a major current challenge to balance the usefulness of collected data and the privacy violation they inevitably cause. Some possible solutions include {\it differential privacy} \cite{dwork2014}, or weaker versions like {\it generative-adversarial privacy} \cite{huang2017}. It could also be possible to combine these with more cryptographic solutions, like {\it homomorphic encryption} or {\it multi-party computation}. It is interesting that such cryptographic solutions may be (essentially) provably robust to any attacker, including a superintelligence\footnote{The possible use of quantum computers may require postquantum cryptography.}.

\section{Dave's world model problem}

Unfortunately, raw data are usually extremely messy, redundant, incomplete, unreliable, poisoning and even hacked. To tackle these issues, it is necessary to infer the likely actual states of the world, given Erin's collected data. This will be Dave's job.

Dave will probably heavily rely on {\it deep representation learning}. This typically corresponds to determining low-dimensional representations of high-dimensional data through deep neural networks. This basic idea has given rise to today's most promising unsupervised machine learning algorithms, e.g. {\it word vectors} \cite{mikolov2013}, {\it autoencoders} \cite{liou2008}, {\it generative adversarial networks} (GANs) \cite{goodfellow2014} and {\it transformers} \cite{VSPUJ+17,RWCLAS2019}.

Given how crucial it is for Dave to have an unbiased representation of the world, much care will be needed to make sure that Dave's inference will foresee selection biases. For instance, when asked to provide images of CEOs, Google Image may return a greater ratio of male CEOs than the actual ratio. More generally, such biases can be regarded as instances of {\it Simpson's paradox} \cite{simpson1951}, and boil down to the saying ``correlation is not causation''. It seems crucial that Dave does not fall into this trap.

In fact, data can be worse than unintentionally misleading. Given how influential Alice may be, there will likely be great incentives for many actors to bias Erin's data gathering to fool Dave. This is known as {\it poisoning attacks}. It seems extremely important that Dave anticipate the fact that the data he was given may be purposely biased, if not hacked. Recent breakthroughs have enabled efficient performant robust high-dimensional statistics \cite{DiakonikolasKane19,LecueDepersin19} with different attack models, as well as applications to robust distributed learning  \cite{blanchard2017,mhamdi2018,damaskinos2018}, though future work will probably need to combine such results with data certification guarantees provided by cryptographical signatures. Eventually, like any good journalist, Dave will likely need to cross information from different sources to infer the most likely states of the world.

This inference approach is well captured by the Bayesian paradigm \cite{hoang2018}. In particular, Bayes rule is designed to infer the likely causes of the observed data $D$. These causes can also be regarded as theories $T$ (and such theories may assume that some of the data were hacked). Bayes rule tells us that the reliability of theory $T$ given data $D$ can be derived formally by $\mathbb P[T|D] = \mathbb P[D|T] \mathbb P[T] / \mathbb P[D]$.

One typical instance of Dave's job is the problem of inferring global health from a wide variety of collected data. This is what has been done by \cite{gbd2016}, using a sophisticated Bayesian model that reconstructed the likely causes of deaths in countries where data were lacking.

Importantly, Bayes rule also tells us that we should not fully believe any single theory. This simply corresponds to saying that data can often be interpreted in many different mutually incompatible manners. It seems important to reason with all possible interpretations rather than isolating a single interpretation that may be flawed.

When the space of possible states of the world is large, which will surely be the case of Dave, it is often computationally intractable to reason with the full posterior distribution $\mathbb P[T|D]$. Bayesian methods often rather propose to approximate it through variational methods, or to sample from the posterior distribution to identify a reasonable number of good interpretations of the data, typically through Markov-Chain Monte-Carlo (MCMC).

In some sense, Dave's job can be regarded as writing a compact report of all likely states of the world, given Erin's collected data. It is unclear what language Dave's report will be in. It might be useful to make it understandable by humans. But this might be too costly as well. Indeed, Dave's report might be billions of pages long. It could be unreasonable or undesirable to make it humanly readable.

Note also that Erin and Dave are likely to gain cognitive capabilities over time. It is surely worthwhile to anticipate the complexification of Erin's data and of Dave's world models. It seems unclear so far how to do so. Some high-level (purely descriptive) language to describe world models may be needed. This high-level language probably should be flexible enough to be reshaped and redesigned over time. This may be dubbed the {\it world description problem}. It is arguably still a very open and uncharted area of research.

\section{Charlie's desirability learning problem}

Given any of Dave's world models, Charlie's job will then be to compute how desirable this world model is. This is the {\it desirability learning} problem \cite{soares2016}, also known as {\it value learning}. This is the problem of assigning desirability scores to different world models. These desirability scores can then serve as the basis for any agent to determine {\it beneficial} actions.

Unfortunately, determining what, say, the median human considers desirable is an extremely difficult problem. But it should be stressed that we should not aim at deriving an {\it ideal} inference of what people desire. This is likely to be a hopeless endeavor. Rather, we should try our best to make sure that Charlie's desirability scores will be {\it good enough} to avoid catastrophic outcomes, e.g. increased polarization, severe misinformation or major discrimination.

One proposed solution is to build upon comparison-based preference inputs of users. For instance, in the Thurstone-Mosteller \cite{thurstone27,mosteller51}, the Bradley-Terry \cite{bradleyterry52} or the Plackett-Luce models \cite{lucebook59,plackett75}, a user is assumed to assign some intrinsic preference $\theta_i$ to each alternative $i$. Given a dilemma between options $i$ and $j$, they will have a probability $\phi(\theta_i-\theta_j)$ to choose $i$ over $j$, where the function $\phi$ is an increasing function from 0 to 1. For instance, we may have $\phi(z) = (1+e^{-z})^{-1}$. Given a dataset of expressed preferences of a user, we may then infer the likely values of $\theta_i$'s.

It is even possible to address the case where the set of alternatives $i \in \mathcal A$ is infinite, as would be the case if alternatives are vectors, or vector representations of alternatives drawn for an unknown large set. One way to do so is to assume a priori that the $\theta_i$'s are a Gaussian process, and that we can define some kernel for the set $\mathcal A$ of alternatives. Bayesian inference (or approximations of Bayesian inference) could then allow to infer the likely preferences of an individual over $\mathcal A$ from a finite sample of binary preferences \cite{chugharamani2005a,chugharamani2005b,chugharamani2005c}.

A perhaps more computationally effective approach could consist in learning a preference vector $\beta_k$ in feature space for each user $k$, so that $\phi(\beta_k^T x_i - \beta_k^T x_j)$ would represent the probability that user $k$ will say to prefer $i$ to $j$ on a given query, where $x_i$ and $x_j$ are vector representations of alternatives $i$ and $j$, as is done in \cite{NGADR18}. By maximizing over $\beta_k$ the likelihood of observed data, with some addditional appropriate regularization term on $\beta_k$, this approach can be argued to compute some maximum a posteriori of user $k$'s preferences.

Another proposed solution to infer human preferences is so-called {\it inverse reinforcement learning} \cite{ng2000,evans2016}. Assuming that humans perform reinforcement learning to choose their actions, and given examples of actions taken by humans in different contexts, inverse reinforcement learning infers what were the humans' likely implicit rewards that motivated their decision-making. Assuming we can somehow separate humans' selfish rewards from altruistic ones, inverse reinforcement learning seems to be a promising first step towards inferring humans' preferences from data.

There are, however, many important considerations to be taken into account, which we discuss below. First, despite Dave's effort and because of Erin's limited and possibly biased data collection, Dave's world model is fundamentally uncertain. In fact, as discussed previously, Dave would probably rather present a distribution of likely world models. Charlie's job should be regarded as a scoring of all such likely world models. In particular, she should not assign a single number to the current state of the world, but, rather, a distribution of likely scores of the current state of the world. This distribution should convey the uncertainty about the actual state of the world. 

Another challenging aspect of Charlie's job will be to provide a useful representation of potential human disagreements about the desirability of different states of the world. Humans' preferences are diverse and may never converge. This should not be swept under the rug. Instead, we need to agree on some way to mitigate disagreement.

This is known as a {\it social choice} problem. In its general form, it is the problem of aggregating the preferences of a group of disagreeing people into a single preference for the whole group that, in some sense, fairly well represents the individuals' preferences. Unfortunately, social choice theory is plagued with impossibility results, e.g. Arrow's theorem \cite{arrow1950} or the Gibbard-Satterthwaite theorem \cite{gibbard1973,satterthwaite1975}. Again, we should not be too demanding regarding the properties of our preference aggregation. Besides, this is the path taken by social choice theory, e.g. by proposing randomized solutions to preserve some desirable properties \cite{hoang2017}.

One particular proposal, known as {\it majority judgment} \cite{balinski2011}, may be of particular interest to us here. Its basic idea is to choose some deciding quantile $q \in [0,1]$ (often taken to be $q=1/2$). Then, for any possible state of the world, consider all individuals' desirability scores for that state. This yields a distribution of humans' preferences for the state of the world. Majority judgment then concludes that the group's score is the quantile $q$ of this distribution. If $q=1/2$, this corresponds to the score chosen by the median individual of the group.


While majority judgment seems promising, it does raise the question of how to compare two different individuals' scores. It is not clear that $score=5$ given by John has a meaning comparable to Jane's $score=5$. In fact, according to a theorem by von Neumann and Morgenstern \cite{neumann1944}, within their framework, utility functions are only defined up to a positive affine transformation. More work is probably needed to determine how to scale different individuals' utility functions appropriately, despite previous attempts in special cases \cite{hoang2016}. 

In practice, another challenge to apply social choice to real-world problem, such as hate speech moderation, is the fact the number of parameters of a hate speech moderation algorithm may be too large for humans to express their preferences. In fact, an algorithm that must decide based on the visual content of a video will likely have to rely on some sophisticated neural network. However, if the hate speech moderation algorithm has $d$ parameters, then the number of options for voters to vote on would be of the order $2^d$. Deciding what should be moderated through classical social choice would then seem impractical.

This problem has been underlined by \cite{MPSW19}, where they studied social choice algorithms with a logarithmic-sized elicitation of voters. In other words, if there were $2^d$ options, then the social choice algorithms could only query $d$ bits of information from each voter. Clearly, we would then obtain suboptimal decision-making. Interestingly, \cite{MPSW19} gave upper and lower bounds on the worst-case ratio between such a social choice algorithm's decision and the optimal decision. Note though that this worst-case ratio may be too conservative, as it excludes reasonable priors we may have on individuals' preferences, or on the similarity of individuals' preferences.

Another fascinating line of research proposed to infer from elicitations of pairwise preferences a model for each user's preferences, and to apply social choice to users' preference models. This is known as {\it virtual democracy} \cite{KLNPP19}. \cite{NGADR18} addressed the case of autonomous car dilemmas, \cite{FBSDC18} faced kidney donation, while \cite{LKKKY+19} tackled food donation. Interestingly, in \cite{LKKKY+19}, voters were given the chance to interact in their model and to receive feedbacks about the social choice algorithm, so as to gain trust in the system that was implemented. This voting-based approach seems to be a promising avenue to alignment.

Now, arguably, humans' current preferences are almost surely undesirable. Indeed, over the last decades, psychology has been showing again and again that human thinking is full of inconsistencies, fallacies and cognitive biases \cite{kahneman2011}. We tend to first have instinctive reactions to stories or facts \cite{bloom2016}, which quickly becomes the position we will want to defend at all costs \cite{haidt2012}. Worse, we are unfortunately largely unaware of why we believe or want what we believe or want. This means that our current preferences are unlikely to be what we would prefer, if we were more informed, thought more deeply, and tried to make sure our preferences were as well-founded as possible.

In fact, arguably, we should prefer what we would prefer to prefer, rather than what we instinctively prefer. Typically, one might prefer to watch a cat video, even though one might prefer to prefer mathematics videos over cat videos. This is sometimes known as humans' {\it volitions} \cite{yudkowsky2004}, as opposed to humans' preferences.


It might even be relevant to consider the volition of much more thoughtful and insightful versions of ourselves, as the currents selves may be blind to some phenomenons that will be of great concerns to our better selves. This can be illustrated by the fact that past standards are often no longer regarded as desirable. Our intuitions about the desirability of slavery, homosexuality and gender discrimination have been completely upset over the last century, if not over the last few decades. It seems unlikely that all of our other intuitions will never change. In particular, it seems unlikely that better versions of ourselves will fully agree with current selves. And it seems reasonable to argue that our better selves would be ``more right'' than current selves.


It is noteworthy that we clearly have epistemic uncertainty about our better selves. Algorithms probably should also take into account their epistemic uncertainty about our better selves' volitions. Computing a given person's volition may be called the {\it volition learning} problem. Interestingly, this is (mostly) a prediction problem, as it is the prediction of the preferences of alternative versions of ourselves. Bayes rule seems able to address this counterfactual reality.

Unfortunately, distinguishing preferences from volitions seems to be a neglected research direction so far. One approach to do so could be to assume that preferences obtained through elicitations better reflect volitions that preferences inferred from inverse reinforcement learning. This could serve as a basis to distinguish humans' behaviors motivated by preferences from those motivated by volitions. Further research in this direction seems critical.

Such scores could also be approximated using a large number of proxies, as is done by {\it boosting methods} \cite{arora2012}. The use of several proxies could avoid the overfitting of any proxy. Typically, rather than relying solely on DALYs \cite{world2009}, we probably should invoke machine learning methods to combine a large number of similar metrics, especially those that aim at describing other desirable economic metrics, like human development index (HDI) or gross national happiness (GNH). Still another approach may consist of analyzing ``typical'' human preferences, e.g. by using {\it collaborative filtering} techniques \cite{ricci2015}.

Computing the desirability of a given world state according to a virtual democracy of our better selves' volitions is Charlie's job. In some sense, Charlie's job would thus be to remove cognitive biases from our intuitive preferences, so that they still basically reflect what we really regard as preferable, but in a more coherent and informed manner. She would then need to aggregate such preferences, which may be computationally challenging if the computations must be done in milliseconds as in the case of autonomous cars \cite{NGADR18} or of recommender systems. 

But she might have to do more. Indeed, it seems critical for us to be able to trust Charlie's computations. But this will arguably be very hard. Indeed, we should expect that our better selves' volitions may find desirable things that our current selves might find repelling. Unfortunately though, we humans tend to react poorly to disagreeing jugments. And this is likely to hold even when the oppositions are our better selves. This poses a great scientific and engineering challenge. How can one be best convinced of the judgments that he or she will eventually embrace but does not yet? In other words, how can we quickly agree with better versions of ourselves? What could be said to an individual to get him closer to his better self? This may be dubbed the {\it individual improvement problem}, which is arguably much more a psychology problem than a computer science one. But educative feedbacks from algorithms may be essential, as seems to have been the case in \cite{LKKKY+19}.



\section{Bob's incentive design}

The last piece of the jigsaw is Bob's job. Bob is in charge of computing the rewards that Alice will receive, based on the work of Erin, Dave and Charlie. Evidently he could simply compute the expectation of Charlie's scores for the likely states of the world. But this is probably a bad idea, as it opens the door to {\it reward hacking} \cite{amodei2016}.

Recall that Alice's goal is to maximize her discounted expected future rewards. But given that Alice knows (or is likely to eventually guess) how her rewards are computed, instead of undertaking the actions that we would want her to, Alice could hack Erin, Dave or Charlie's computations, so that such hacked computations yield large rewards. This is sometimes called the {\it wireheading problem} \cite{EverittHutter19}.


To avoid wireheading, Bob's role will be to make sure that, while Alice's rewards do correlate with Charlie's scores, they also give Alice the incentives to guarantee that Erin, Dave and Charlie perform as reliably as possible the job they were given.

In fact, it even seems desirable that Alice be incentivized to constantly upgrade Erin, Dave and Charlie for the better. After all, it is high unlikely that early versions of Erin, Dave and Charlie will have no flaws. In fact, because of Goodhart's law \cite{ManheimGarrabrant18}, even the slight imperfection in their design might result in highly counter-productive optimization by Alice. Indeed, \cite{EHR20} showed that when the discrepancy between a measure and the true goal has a heavy-tailed distribution, then as the measure gets maximized, the correlation between measure and goal becomes negative.

This suggests that {\it robust alignment} must rely on the continuous upgrade of the reward system, so that the computation of Alice's rewards keep approaching the objective function that we really want to maximize. This is sometimes known as {\it corrigibility} \cite{SFAY15,LWN19}, though for efficient corrections, it seems essential that this corrigibility be pre-programmed. Bob's job will be to solve the {\it programmed corrigibility} problem, by incentivizing Alice to upgrade adequately the components of her reward system.


Unfortunately, it seems unclear how Bob can best make sure that Alice has such incentives. Perhaps a good idea is to penalize Dave's reported uncertainty about the likely states of the world. Typically, Bob should make sure Alice's rewards are affected by the reliability of Erin's data. The more reliable Erin's data, the larger Alice's rewards.  Similarly, when Dave or Charlie feel that their computations are unreliable, Bob should take note of this and adjust Alice's rewards accordingly to motivate Alice to provide larger resources for Charlie's computations. All of this seems to demand a way for Bob to estimate the performance of the different components of the reward system. Doing so seems to be an open problem so far.

Now, Bob should also mitigate the desire for more reliable data and for more trustworthy computations with the fact that such efforts will necessarily require the exploitation of more resources, probably at the expense of Charlie's scores. It is this non-trivial trade-off that Bob will need to take care of.

Bob's work might be simplified by some (partial) control of Alice's action or world model. Although it seems unclear so far how, techniques like {\it interactive proofs} (IP) \cite{babai1985,goldwasser1989} or {\it probabilistically checkable proofs} (PCP) \cite{arora1998} might be useful to force Alice to prove its correct behavior. By requesting such proofs to yield large rewards, Bob might be able to incentivize Alice's transparency. All such considerations make up Bob's {\it incentive problem}.



\section{Decentralization}

We have decomposed robust alignment into 5 components for the sake of exposition. However, any component will likely have to be decentralized to gain reliability and scalability. In other words, instead of having a single Alice, a single Bob, a single Charlie, a single Dave and a single Erin, it seems crucial to construct multiple Alices, Bobs, Charlies, Daves and Erins.

This is key to {\it crash-tolerance}. Indeed, a single computer doing Bob's job could crash and leave Alice without reward nor penalty. But if Alice's rewards are an aggregate of rewards given by a large number of Bobs, then even if some Bobs crash, Alice's rewards will remain mostly the same. But crash-tolerance is likely to be insufficient. Instead, we should design {\it Byzantine-resilient} mechanisms, that is, mechanisms that still perform correctly despite hacked or malicious Bobs. Robust estimators \cite{DiakonikolasKane19,LecueDepersin19} for both distributed workers \cite{blanchard2017} and parameter servers \cite{EGGR19} may be useful for this purpose.

\begin{figure*}[h!]
\includegraphics[width=\textwidth]{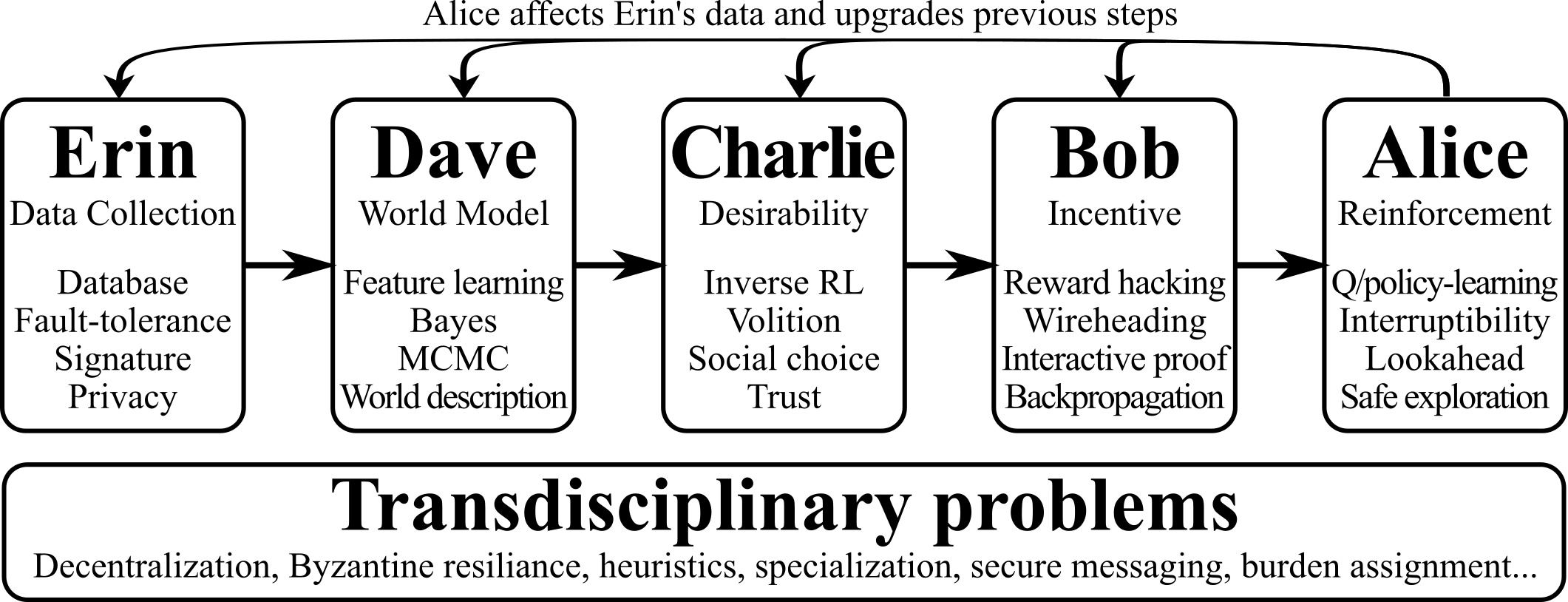}
\caption{We decompose robust alignment into 5 steps. Each step is associated with further substeps or techniques. Also, there are critical subproblems that will likely be useful for several of the 5 steps.}
\label{fig:complete_roadmap}
\end{figure*}

Evidently, in this Byzantine environment, cryptography, especially (postquantum?) cryptographical signatures and hashes, are likely to play a critical role. Typically, Bobs' rewards will likely need to be signed. More generally, the careful design of secure communication channels between the components of the algorithms seems key. This may be called the {\it secure messaging problem}.

Another difficulty is the addition of more powerful and precise Bobs, Charlies, Daves and Erins to the pipeline. It is not yet clear how to best integrate reliable new comers, especially given that such new comers may be malicious. In fact, they may want to first act benevolent to gain admission. But once they are numerous enough, they could take over the pipeline and, say, feed Alice with infinite rewards. This is the {\it upgrade problem}, which was recently discussed by \cite{christiano2018} who proposed using numerous weaker algorithms to supervise stronger algorithms. More research in this direction is probably needed.

Now, in addition to reliability, decentralization also enables different Alices, Bobs, Charlies, Daves and Erins to focus on specific tasks. This would allow for more optimized solutions at lower costs. To this end, it may be relevant to adapt different Alices' rewards to their specific tasks. Note though that this could also be a problem, as Alices may enter in competition with one another like in the prisoner's dilemma. We may call it the {\it specialization problem}. 


Another open question is the extent to which algorithms should be exposed to Bobs' rewards. Typically, if a small company creates its own algorithm, to what extent should this algorithm be aligned? It should be noted that this may be computationally very costly, as it may be hard to separate the signal of interest to the algorithm from the noise of Bobs' rewards. Intuitively, the more influential an algorithm is, the more it should be influenced by Bobs' rewards. But even if this algorithm is small, it may be important to demand that it be influenced by Bobs to avoid any {\it diffusion of responsibility}, i.e. many small algorithms that disregard safety concerns on the ground that they each hardly have any global impact on the world.

What makes this nontrivial is that any algorithm may gain capability and influence over time. An unaligned weak algorithm could eventually become an unaligned human-level algorithm. To avoid this, even basic, but potentially unboundedly self-improving\footnote{In particular, nonparametric algorithms should perhaps be treated differently from parametric ones.} algorithms should perhaps be given at least a seed of alignment, which may grow as algorithms become more powerful. More generally, algorithms should strike a balance between some original (possibly unaligned) objective and the importance they give to alignment. This may be called the {\it alignment burden assignment problem}.

Figure \ref{fig:complete_roadmap} recapitulates our complete roadmap.







\section{Conclusion}

This paper discussed the {\it robust alignment} problem, that is, the problem of aligning the goals of algorithms with human preferences. It presented a general roadmap to tackle this issue. Interestingly, this roadmap identifies 5 critical steps, as well as many relevant aspects of these 5 steps. In other words, we have presented a large number of hopefully more tractable subproblems that readers are highly encouraged to tackle. Hopefully, this combination allows to better highlight the most pressing problems, how every expertise can be best used to, and how combining the solutions to subproblems might add up to solve robust alignment.

\bibliography{alignment}
\bibliographystyle{plain}

\end{document}